# PERCORE: A Deep Learning-Based Framework for Persian Spelling Correction with Phonetic Analysis

Seyed Mohammad Sadegh Dashti[1], Amid Khatibi Bardsiri[1], Mehdi Jafari Shahbazzadeh[2]

[1] Computer Engineering Department, Kerman Branch, Islamic Azad University, Kerman, Iran

[2] Electrical Engineering Department, Kerman Branch, Islamic Azad University, Kerman, Iran.

Corresponding Author: Amid Khatibi Bardsiri; a.khatibi@srbiau.ac.ir

## Abstract

This research introduces a state-of-the-art Persian spelling correction system that seamlessly integrates deep learning techniques with phonetic analysis, significantly enhancing the accuracy and efficiency of natural language processing (NLP) for Persian. Utilizing a fine-tuned language representation model, our methodology effectively combines deep contextual analysis with phonetic insights, adeptly correcting both non-word and real-word spelling errors. This strategy proves particularly effective in tackling the unique complexities of Persian spelling, including its elaborate morphology and the challenge of homophony. A thorough evaluation on a wide-ranging dataset confirms our system's superior performance compared to existing methods, with impressive F1-Scores of 0.890 for detecting real-word errors and 0.905 for correcting them. Additionally, the system demonstrates a strong capability in non-word error correction, achieving an F1-Score of 0.891. These results illustrate the significant benefits of incorporating phonetic insights into deep learning models for spelling correction. Our contributions not only advance Persian language processing by providing a versatile solution for a variety of NLP applications but also pave the way for future research in the field, emphasizing the critical role of phonetic analysis in developing effective spelling correction system.

## 1. Introduction

Spelling correction is an integral component of all text processing environments, particularly for languages with complex morphology and syntax like Persian. The task of spelling correction primarily involves the detection and correction of two types of errors: non-word errors and real-word errors.

Non-word errors are instances where the misspelled words are not found in a dictionary and are meaningless. These errors often occur due to typographical mistakes or a lack of knowledge about the correct spelling of a word. While non-word errors in Persian are mostly similar to those in other languages, there are unique cases due to the specific features of the Persian language. For instance, consider the word ("می‌نوازم" /minævɑzæm/ 'I play')[1]. If a user ignores the pseudo space, the word transforms to "مینوازم", which might not be found in a standard dictionary. Conversely, if the user commits a split error and uses a white space instead of a pseudo space, then the word transforms into two different sequences of characters — "می" and "نوازم" — neither of which can be attested in the dictionary.

Real-word errors in spelling correction are a significant challenge. They occur when a correctly spelled word is used incorrectly in context. These errors can be due to accidental mistyping, confusion between similar sounding or meaning words [1], incorrect replacements by automated systems like AutoCorrect features [2], and misinterpretation of input by Automatic Speech Recognition (ASR) and Optical Character Recognition (OCR) systems [3-6].

The Persian language, with its extensive vocabulary and intricate properties, further complicates real-word error correction. Unique features of Persian such as homophony (words that sound the same but have different meanings), polysemy (words with multiple meanings), heterography (words with the same spelling but different meanings depending on pronunciation), and word boundary issues add to this complexity.

Despite these inherent challenges, there have been concerted efforts to develop both statistical and rule-based methods for detecting and correcting both classes of errors in Persian. However, these methods have achieved only limited success. The task of word error detection and correction in Persian continues to be an active area of research. The advent of more advanced methods and tools holds promise for overcoming these challenges and enhancing the accuracy of error detection and correction. The objective of this study is to introduce an advanced method for detecting and correcting spelling errors in Persian. Our primary contributions are summarized as follows:

- **General Persian Corpus:** We introduce a large general corpus of Persian text, consisting of 1.4 million formal documents.
- **Language Representation Model:** We present a language representation model that has been fine-tuned for the task of spelling correction. Our method utilizes the contextual score of words to determine the best replacement candidate for an error.
- **Error Generation Algorithm:** We propose a novel error generation algorithm for populating original corpora of Persian text with artificially generated errors. This algorithm offers flexibility in specifying the proportion of real-word and non-word errors in a given corpus. Importantly, it can also determine the edit distance of generated errors, allowing NLP researchers to set up various test scenarios tailored to their specific research questions.

[1] All pronunciations have been provided in International Phonetic Alphabet (IPA)

- **Persian Soundex:** We introduce a Persian Soundex algorithm that measures the phonetic similarity between pairs of words. This algorithm has proven useful in word error correction in other languages. We adapt this algorithm based on the Persian alphabet and vocabulary and use it to improve error correction accuracy.

We assess the effectiveness of our hybrid approach using evaluation metrics such as F1-score and compare our approach to existing techniques for non-word and real-word error detection and correction.

The remainder of this paper is structured as follows: We begin with a review of prior research in the field. Next, we explain the challenges faced in Persian language text processing. Subsequently, we describe our proposed approach. Evaluation and experiment results are then presented and discussed. In the final segment, we conclude our findings.

## 2. Related Works

Automatic word error correction is a pivotal component in Natural Language Processing (NLP) systems. Initial techniques were reliant on edit distance and phonetic algorithms [7-10], with subsequent enhancements demonstrating the effectiveness of incorporating context information to boost the efficiency of auto-correction systems [11]. Utilization of contextual measures such as semantic distance and noisy channel models based on N-grams has been widespread across various NLP applications [1, 2, 12-14]. Innovative strategies have been devised to correct multiple real-word errors in highly noisy contexts [15], including a notable model by Dashti for detecting and auto-correcting real-word errors within sequences containing multiple inaccuracies [16].

The adoption of neural word or sense embeddings, leveraging context information for spelling correction, marked a significant advancement [17]. The use of pretrained contextual embeddings for detecting and correcting real-word errors was a notable milestone [18]. By 2020, deep learning methodologies were applied to address context-sensitive spelling errors in English documents [19], with subsequent research developing a BERT-based model for similar purposes [20].

The rapid progression and innovation in pre-trained language models, exemplified by the development of BERT and the subsequent introduction of the GPT series—including GPT-2.0, GPT-3.0, have markedly revolutionized the field of error correction. These advancements have ushered in an era of unparalleled adaptability and enhanced performance across a diverse array of languages, as documented in seminal works [21, 22]. Notably, the application of these models to English and Arabic has demonstrated their proficiency in leveraging vast knowledge bases for the effective correction of complex error types through fine-tuned, prompt-based methods [23, 24]. In parallel, the exploration and utilization of large language models for Chinese language processing, as seen with innovations like SpellBERT and methodologies that employ disentangled phonetic representations, highlight the bespoke adaptations necessary to address the unique challenges presented by Chinese orthography and phonology [25, 26]. Further contributions to the Chinese language processing field have introduced phonetic pre-training techniques [27], enhancing the model's linguistic comprehension. Additionally, NeuSpell emerges as a neural spelling correction toolkit, offering a suite of pre-trained models designed for straightforward, user-friendly implementation [28].

In clinical text, Tran et al. introduced a context-sensitive spelling correction model [29], and a contextual spelling correction approach tailored for end-to-end speech recognition systems was developed [30]. A multi-task detector-corrector framework for Chinese spelling correction was also proposed [31]. Liu et al. advanced Chinese spelling correction with CRASpell, a contextual typo robust approach [32]. Furthermore, AraSpell employed a Transformer model for Arabic spelling correction, learning the relationships between words and their misspellings [33].

This body of research collectively underscores a significant leap forward in the development of sophisticated, nuanced error correction tools capable of navigating the complexities inherent in multiple languages.

Despite these advancements, the exploration of sophisticated models like BERT and GPT for Persian language spelling correction remains limited. Early efforts in Persian spelling correction have relied on statistical or rule-based methods, with systems like Vafa spellchecker and the work of Mosavi and Miangah using N-grams, a monolingual corpus, and string distance measures to tackle spelling challenges in Persian [34-41]. The transition from foundational models to advanced AI technologies such as BERT and GPT highlights a promising area, aiming to leverage these models to address the unique challenges of Persian spelling correction.

### 2.1 Linguistic Challenges in Persian Automatic Spelling Correction

Persian, an integral member of the Indo-Iranian group within the Indo-European language family, serves as the official language in Iran, Tajikistan, and Afghanistan. Its deep historical roots contribute to a rich linguistic tapestry. This language, enriched by Arabic elements yet retaining its core structure over centuries, poses unique challenges in the field of Automatic Spelling Correction (ASC) [39, 42]. The complexity of Persian, especially apparent in its script and morphology, stands in stark contrast not only to the Latin alphabet of English, with its irregular phoneme-grapheme correspondences [43], but also to the cursive script of Arabic, which grapples with complexities in its system of vowelization [44].

Adding to this linguistic diversity, Chinese introduces a set of challenges distinct from those of Persian, English, and Arabic. As a language with thousands of years of history, it plays a central role in East Asia, being the official language of China, Taiwan, and one of the official languages of Singapore [45]. Chinese is characterized by its logographic writing system, where each character represents a morpheme and can be a word on its own or part of a compound word [46]. This system is fundamentally different from the alphabetic and abjad systems used by English, Persian, and Arabic, respectively. The challenges in ASC for Chinese stem primarily from this logographic nature, the high degree of homophony due to its tonal system, and the absence of a clear boundary between words in written text [47]. These characteristics necessitate ASC approaches that are highly sensitive to the context and semantic content of text, as well as sophisticated algorithms for character recognition and word segmentation.

Thus, the diversity in linguistic structures across these languages—ranging from the complexity of scripts and morphology in Persian, the irregular phoneme-grapheme correspondences in English, the vowelization system in Arabic, to the logographic writing system and tonality of Chinese—underscores the necessity for ASC systems to be finely attuned to the phonetic, orthographic, and grammatical nuances of each language. This multifaceted landscape presents a compelling challenge for developing effective and nuanced ASC technologies that can accommodate the rich variety of human language.

Incorporating Chinese into the comparative analysis of linguistic challenges in Automatic Spelling Correction (ASC) alongside Persian, English, and Arabic enriches our understanding of the unique and shared hurdles each language presents. This broader comparison underscores the diverse nature of linguistic structures and the necessity for ASC systems to be highly tailored to address the specific challenges of each language:

- **Writing System Complexity:** Persian employs an Arabic-derived script with additional characters and nuanced diacritics, complicating its script beyond the straightforward Latin alphabet of English or the cursive, vowelization-dependent script of Arabic. Chinese, distinctively, uses a logographic system where characters represent morphemes or whole words, introducing challenges in character recognition and segmentation not found in phonetic or alphabetic systems.
- **Homophonic Variability:** Persian and Chinese both deal with extensive homophony; Persian's arises from its etymological diversity, while Chinese's stems from its tonal nature and limited syllable inventory. English also faces homophonic challenges due to its irregular spellings. Arabic, while phonetically transparent, is not without its homophonic issues, albeit to a moderate extent.
- **Vowel Representation:** Persian's context-dependent vowel representation contrasts sharply with English's relatively stable vowel system and Arabic's tendency to omit short vowels in informal writing. Chinese stands apart by not employing a vowel representation system per se but instead uses tones to distinguish meanings across its syllable-based script.
- **Morphological Density:** Persian and Arabic exhibit a complex morphological structure, with Persian involving intricate affixation and Arabic featuring a root-pattern morphology. English presents a less complex morphology. Chinese, while having a simpler morphological structure regarding affixation, demonstrates complexity in compound word formation, necessitating effective segmentation strategies.
- **Diacritic Variability:** Persian and Arabic experience variability and omission of diacritics, respectively, which can impact word recognition. English diacritics are consistent across dialects. Chinese, using tone marks in its Pinyin romanization system, faces challenges in tone recognition and representation, though these are not used in the standard Chinese script.

Reflecting these insights, the extended comparison in Table 1, highlights the nuanced challenges ASC systems face in adapting to the linguistic features of Persian, English, Arabic, and now Chinese:

Table 1: Comparative Overview of Linguistic Challenges in ASC for Persian, English, Arabic and Chinese

| Feature | Persian ASC Challenges | English ASC Challenges | Arabic ASC Challenges | Chinese ASC Challenges |
|---|---|---|---|---|
| Writing System | Arabic script with additional characters and diacritics | Latin alphabet with phoneme-grapheme irregularities | Cursive script with vowelization complexities | Logographic system, representing morphemes/words |

| Feature | Persian ASC Challenges | English ASC Challenges | Arabic ASC Challenges | Chinese ASC Challenges |
|---|---|---|---|---|
| Homophones | Extensive, due to etymological diversity | Significant, from irregular spellings | Moderate, with phonetic transparency | High, due to tonal language and limited syllable inventory |
| Vowel Representation | Context-dependent, non-unique representation | Relatively stable but occasionally irregular | Short vowels often omitted in informal writing | N/A, uses tones for different meanings |
| Morphological Density | Complex affixation and word formation | Less complex morphology | Root-pattern morphology | Simple in affixation, complex in compound word formation |
| Diacritic Variation | Variable by dialect and style | Consistent across dialects | Consistent but often omitted | Tone marks crucial in Pinyin, not used in standard script |

This comprehensive overview illustrates the diverse requirements for ASC systems across different languages, emphasizing the importance of specialized approaches to effectively address the unique combination of script, phonology, morphology, and orthographic practices in Persian, English, Arabic, and Chinese.

On the other hand, Persian's richness and expressiveness as a language, however, bring forth numerous challenges in its processing, particularly in the realm of Automatic Spelling Correction (ASC):

- **Character Ambiguity**: Persian characters like "ی" and "ي" are often used interchangeably but represent different sounds [48].
- **Rich Morphology** New words can be created by adding prefixes and suffixes to a base word, like "دست" (hand) to "دست‌ها" (hands) [49].
- **Orthography**: Persian involves a combination of spaces and semi-spaces, which can lead to inconsistencies [50].
- **Co-articulation**: The pronunciation of a consonant like "ب" can be affected by the subsequent vowel. [51].
- **Dialectal Variation**: Persian has several standard varieties such as Farsi, Dari, and Tajik [52].
- **Cultural Factors**: The phenomenon of persianization can shape the way Persian is used and interpreted.
- **Lack of Resources**: Persian is often categorized as a low-resource language due to the limited availability of data and tools for natural language processing [49].
- **Free Word Order**: Persian allows for the rearrangement of words within a sentence without significantly altering its meaning [53].
- **Homophony**: Different words have identical pronunciation but different meanings, like ("گذار" /gʊzɑr/ 'transition') and ("گزار" /gʊzɑr/ 'predicate') [54].

- **Diacritics**: They are frequently left out in writing, leading to ambiguity in word recognition [55].
- **Rapidly Changing Vocabulary**: Persian’s vocabulary is rapidly evolving due to factors such as technology, globalization [56].
- **Lack of standardization**: There isn’t a single standard for Persian text, which can complicate the development of language processing models capable of handling a variety of dialects and styles [57].

## 3. Material and Methods

Our method detects and corrects two types of errors in Persian text: Non-word and Real-word errors. Figure 1 demonstrates the architecture of the proposed system. The system design consists of six distinct modules that exchange information through a databus. The INPUT module takes raw test corpora. The pre-processing component normalizes the text and handles word boundary issues. The error generation module generates errors based on a desired density at different edit-distance function values. The contextual analyzer module evaluates the contextual similarity within desired word sequences. Error detection uses a dictionary look-up method for non-word error detection and contextual similarity matching for real-word errors. The error correction module corrects both classes of errors using context information from fine-tuned contextual embeddings model, along with phonetic and edit-distance similarity measures. The corrected corpora or word sequence is delivered through the OUTPUT module.

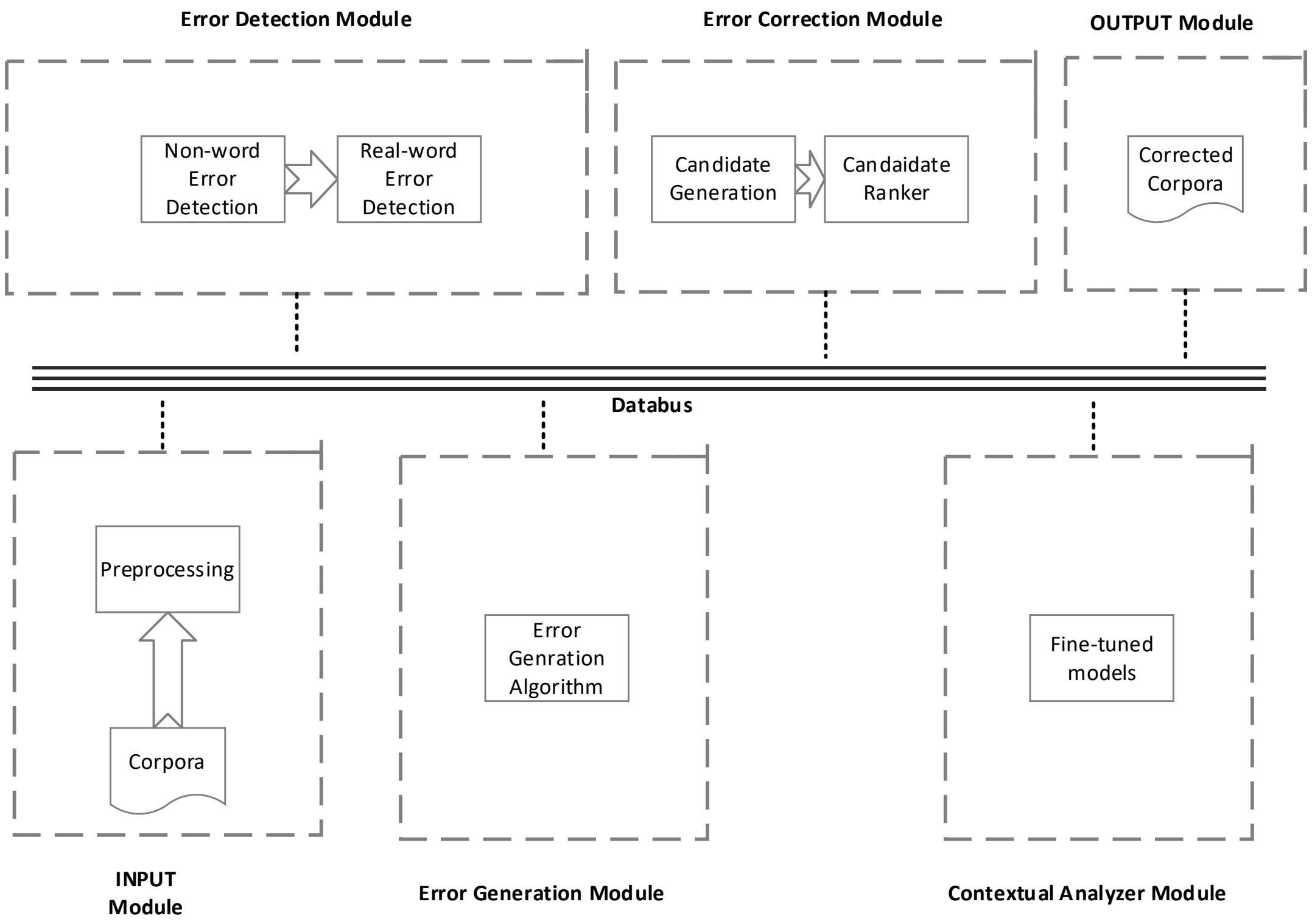

Figure 1. Architecture of the Proposed System for Generating, Detecting and Correcting Persian Word Errors.

### 3.1 Pre-processing step

Text pre-processing is a crucial step in most NLP applications, consisting of sentence segmentation, tokenization, normalization, and stop-word elimination. Sentence segmentation involves determining the boundaries of sentences, typically divided by punctuation marks such as periods, exclamation points, or question marks. Tokenization is the process of dividing a sentence into a sequence of terms that represent the sentence, which will be used to extract features. Normalization involves converting text into canonical forms and is an important step in Persian NLP applications, as it is in many other languages.

One of the key tasks in normalizing Persian text is converting pseudo and white spaces to regular forms, substituting whitespaces with zero width nonjoiners when necessary. For example, ('می بینم' /mibinæm/ 'see') is replaced with ('می‌بینم' / mibinæm / 'see'). Persian and Arabic share many characteristics, and some Persian letters are often misspelled using Arabic forms. Researchers often find it beneficial to standardize these variations by replacing Arabic characters (ي 'Y' /j/; ك 'k' /k/; ة 'h' /h/) with their Persian equivalents. For instance, ('راي' /ræy/ 'vote') is transformed to ('رای' /ræy/ 'vote'). Normalization also includes removing diacritics from Persian words; e.g., ('درّه' /dærre/ 'valley') is changed to ('دره' /dære/ 'valley'). Additionally, Kashida(s) are removed from words; for instance, ('بـــانک' /bɑnk/ 'bank') is transformed to ('بانک' /bɑnk/ 'bank'). To achieve the normalization goal, a dictionary including the correct typographic form of all Persian words named Dehkhoda is used to find the normal form of multi-shaped words [58].

### 3.2 Damerau-Levenshtein distance

Our approach uses the Damerau-Levenshtein distance measure to generate non-word and real-word errors for detection and correction tasks. This measure considers insertion, deletion, substitution, and transposition of characters. For example, the Damerau-Levenshtein distance between "KC" and "CKE" is 2. It's found that around 80% of human-generated spelling errors involve these four error types [59]. Studies show that real-word errors account for about 25% to 40% of all spelling errors in English text [60, 61]. We choose an edit distance of up to 2 between the correct term and the error. When the edit distance is set to one, an average of five candidates are generated as replacements for a target context word. However, once the distance is increased to 2, the average number rises to 23. The computing time increases correspondingly. We ensure that the generated candidates are attested in the reference lexicon.

### 3.3 Error Generation

In the following two sections, we describe the error generation algorithm and its properties, as well as explain how we determine the density of errors.

#### 3.3.1 Error Density

There are no publicly available Persian texts containing genuine word errors with reasonable density. Researchers address this issue by inserting artificially generated misspellings into free-running text. For instance, the authors of [62] randomly generated malapropisms into the test corpus in free-running English texts at a frequency rate of 0.005, replacing 1 word in every 200 words. Different frequency rates were used in [1, 15, 16]. The authors used an approach based on confusion sets for the Persian language [34, 63], although they did not specify the real-word error density in the test corpus. However, based on the size of the evaluation corpora and the number of test instances, it can be estimated that it amounts to about 15% of total sentences. In the evaluation of spelling correction systems, there is a unique constraint where the maximum number of errors per sentence is set to one when generating word errors, meaning that each distinct sentence may contain only one error. In our proposed model, we follow this strategy and generate only one word error per sentence. However, we test various error density rates, ranging from 10% to 50% of the sentences in the test corpus containing a word error, where 10 represents a normal source of error and 50 represents a very noisy error source. This high density of errors was used to assess the effectiveness of our proposed technique.

### 3.3.2 Error Generation Algorithm

We proposed a novel error generation algorithm for populating pre-processed corpora. Pseudocode1 is used to generate errors in a corpus that is essentially free of misspellings. The method introduces artificially generated errors into the original test corpora within distances 1 and 2. The number of sentences with errors is determined by the error density $E$. For each target term in the text, we extract all probable occurrences of error that are within the appropriate distance in the dictionary. To produce errors and replacement candidates, the suggested model uses a large lexicon borrowed from the Vafa spell-checker for the Persian language dictionary, which contains 1,095,959 different terms. In the error generation process, we adhere to a strict rule: only those original corpora words that can be attested in the lexicon may be replaced with an instance of error, in order to avoid any Out-Of-Vocabularies(OOV).

Furthermore, parameter $D1$ allows us to determine the number of errors within distance 1, with the remaining errors at distance 2 denoted as $1 - D1$. For example, if there is a 1000-sentence test corpus with $E$ = 0.20 and $D1$ = 0.60, then 200 sentences will contain misspellings, with 120 of them having distance 1 misspellings and 80 of them including errors within distance 2. Following the observations of [8] and [59] and keeping in mind the objective of the model, we classify word errors into two types and assign a specific density to each one. β represents the total number of artificially generated errors in the erroneous corpora, while $\beta_1$ and $\beta_2$ are coefficient parameters representing the density of non-word and real-word errors, respectively. Equation 1 represents the relationship between the size of the corpora and the frequency of non-word and real-word errors.

$$\beta \begin{cases} N * E \\ \beta_1 + \beta_2 \leq 1 \end{cases} \qquad (1)$$

Our novel error generation algorithm allows us to set up various scenarios and evaluate the robustness of our model in detecting and correcting real-word and non-word errors at the desired density and complexity.

Pseudocode1: Error Generation Algorithm

| | |
|---|---|
| **Input:** | Persian text corpus as $P$; Density of Error as $E$, Errors within distance 1 as $D1$; Errors within distance 2 as $1-D1$;<br><br>non-word error density at one unit distance as $\beta_1 L_1$;<br>non-word error density at two-unit distance as $\beta_1 L_2$;<br><br>real-word error density at one unit distance as $\beta_2 L_1$;<br>real-word error density at two-unit distance as $\beta_2 L_2$;<br><br>$0.1 \le E \le 0.50$; $0 \le D1 \le 1$; $\beta_1 + \beta_2 \le 1$ |
| **Output:** | Test corpus $EP$ with generated errors included in it |
| **Start:** | |
| **1** | Initialize parameters $E, D1$, $\beta_1$, $\beta_2$ |
| **2** | Recognize all sentences $s_1, s_{2,\ldots,}s_N$ boundaries |
| **3** | Select $N{*}\,E$ sentences randomly; as random list $RLT$ |
| | *## $RLT_1$ is the list of sentences where errors will be within edit-distance 1.* |
| **4** | Create Random list $RLT_1$; where ($RLT_1 \subseteq RLT$) AND ($RLT_1 = RLT * D1$) |
| | *## $RL_2$ is the list of sentences where errors will be within edit-distance 2.* |
| **5** | Create Random list $RLT_2$; where ($RLT_2 \subseteq RLT - RLT_1$) AND ($RLT_2 = RLT * (1 - D1)$) |
| | *## Replacing target words with generated errors of different type* |
| **6** | For existing $SL_i s \in RLT_1$ do:<br>Create Random lists with following details:<br>($\beta_1 L_1 * count(RLT_1)$ as $RLT_1\beta_1$); ($\beta_2 L_1 * count(RLT_1)$ as $RLT_1\beta_2$); |
| **7** | For existing $SL_i s \in RLT_2$ do:<br>Create Random lists with following details:<br>($\beta_1 L_2 * count(RLT_2)$ as $RLT_2\beta_1$); ($\beta_2 L_2 * count(RLT_2)$ as $RLT_2\beta_2$); |
| **8** | For existing $SL_i s \in \left(RLT_1\beta_2 \; and \; RLT_1\beta_1\right) \; and \; SL_i s \in (RLT_2\beta_1 \; and \; RLT_2\beta_2)$ do:<br><br>Select word $w_n$ randomly from $SL_i$<br><br>Replace $w_n$ with an instance $w'_n$ of related error set $E'$; with distance 1 if $SL_i \in \left(RLT_1\beta_2 \; and \; RLT_1\beta_1\right)$ OR with distance 2 if $SL_i \in (RLT_2\beta_1 \; and \; RLT_2\beta_2)$ |

## 3.4 Contextual embeddings

Word embeddings analyze large volumes of textual data to embed word meanings into low-dimensional vectors [64, 65]. They store valuable syntactic and semantic information [66] and are beneficial to many NLP applications [67]. However, they suffer from meaning conflation deficiency: the inability to distinguish between a word's multiple meanings.

To address this, state-of-the-art approaches represent specific word senses, referred to as contextual embeddings or sense representation. Contextualized word embedding methods like ELMo consider the context of the input sequence [64]. There are two ways to pre-train language representations: feature-based approaches like ELMo and fine-tuning approaches like OpenAI GPT [22]. Fine-tuning methods train a language model using massive datasets of unlabeled plain texts. The parameters of these models are then fine-tuned using task-specific data [22, 68, 69].

However, pre-training an efficient language model requires significant data and computational resources [70-73]. Multilingual models have been developed for languages with identical morphology and syntactic organization. But non-Latin languages differ greatly from Latin-based languages and require a language-specific approach [74]. A similar issue exists in the Persian language. Despite the fact that some multilingual models include Persian, they may fall behind monolingual models that are specifically trained on language-specific vocabulary with larger volumes of Persian text data. To the best of our knowledge, ParsBert [75] is the only attempt to pre-train a Bidirectional Encoder Representation Transformer (BERT) [68] model for the Persian language.

### 3.4.1 PERCORE's Language Representation Model

Persian is considered an under-resourced language. Despite the existence of language models that support Persian, only one has been pre-trained on a large Persian corpus [75]. However, ParsBERT's data includes many informal documents, such as user reviews and comments, and many of the collected documents contain misspelled words, making the model unsuitable for spelling correction tasks. The absence of a high-performance language model in this field is a significant challenge. In this section, we will discuss our General Persian Corpus and the process of pre-training PERCORE's language representation model.

### 3.4.2 Data

While there are many freely available formal Persian texts, they have not been compiled into a single, error-free, large corpus. This is necessary for the purpose of pre-training a language representation model for spelling correction. As a result, to train an effective model for spelling correction in Persian, we had to gather a large collection of texts from available corpora of formal text. This corpus contains 1.4 million documents collected from various sources:

- **Bijankhan Corpus:** The Bijankhan corpus[2] is a collection of tagged texts, including daily news and common texts [76]. It comprises 4300 articles that have been categorized into different subject categories such as political, cultural, and others. With around 2.6 million words, this corpus provides a rich source of data for researchers working on Persian language processing

[2] https://dbrg.ut.ac.ir/%d8%a8%db%8c%da%98%d9%86%e2%80%8c%d8%ae%d8%a7%d9%86/

- **Hamshahri Corpus:** The second version of the Hamshahri corpus[3] was released on October 20, 2008 [77]. It contains 323,616 text stories organized into 82 categories and 65 subjects, with an average article length of 380 words.
- **Persian Wikipedia Corpus:** The Persian Wikipedia corpus[4] is a collection of 441 million tokens in Persian text, which comprises 1,160,676 articles from the Persian Wikipedia as of October 1, 2018.

The collected documents undergo normalization, pre-processing, and cleaning to remove elements such as POS tags, HTML tags, hyperlinks, etc.

### 3.4.3 Model Architecture

PERCORE's architecture uses the original $\mathbf{BERT_{BASE}}$ configuration with 12 hidden layers, 12 attention heads, 768 hidden sizes, and a total of 110M parameters. Our model has a maximum token capacity of 512. Model's architecture is shown in Figure 2.

Many attribute BERT's success to its MLM pre-training task , where it masks or replaces tokens at random before predicting the original tokens [68]. This makes BERT well-suited to be a spelling checker, as it interprets the masked and changed tokens as misspellings. In BERT's embedding layer, each input token $\mathrm{T_i}$ is indexed to its embedding representation $\mathrm{ER_i}$. Then, $\mathrm{ER_i}$ is forwarded to BERT's encoder layers to obtain the following representation of $\mathrm{HR_i}$:

$$\mathrm{ER_i} = \mathrm{BERT - Embedding(T_i)} \qquad (2)$$
$$\mathrm{HR_i} = \mathrm{BERT - Encoder(ER_i)} \qquad (3)$$

where $\mathrm{ER_i}$ and $\mathrm{HR_i} \in R^{1*d}$ and $d$ is the hidden dimension. Following this, similarities between $\mathrm{HR_i}$ and all token embeddings will be estimated to predict the distribution of $\mathrm{Y_i}$ over the existing vocabulary.

$$\mathrm{Y_i} = \mathrm{Softmax}(\mathrm{HR_i}, \boldsymbol{E}^T) \qquad (4)$$

where $\boldsymbol{E} \in R^{V*d}$ and $\mathrm{Y_i} \in R^{1*V}$; here $V$ represents the size of the vocabulary and $\boldsymbol{E}$ denotes the BERT embedding layer. The $i$th row of $\boldsymbol{E}$ corresponds to $\mathrm{ER_i}$ in Equation 2. The final rectification result for $\mathrm{T_i}$ is the $\mathrm{T_k}$ token, whose $\mathrm{ER_k}$ has the highest similarity to $\mathrm{HR_i}$.

[3] https://dbrg.ut.ac.ir/hamshahri/
[4] https://github.com/Text-Mining/Persian-Wikipedia-Corpus

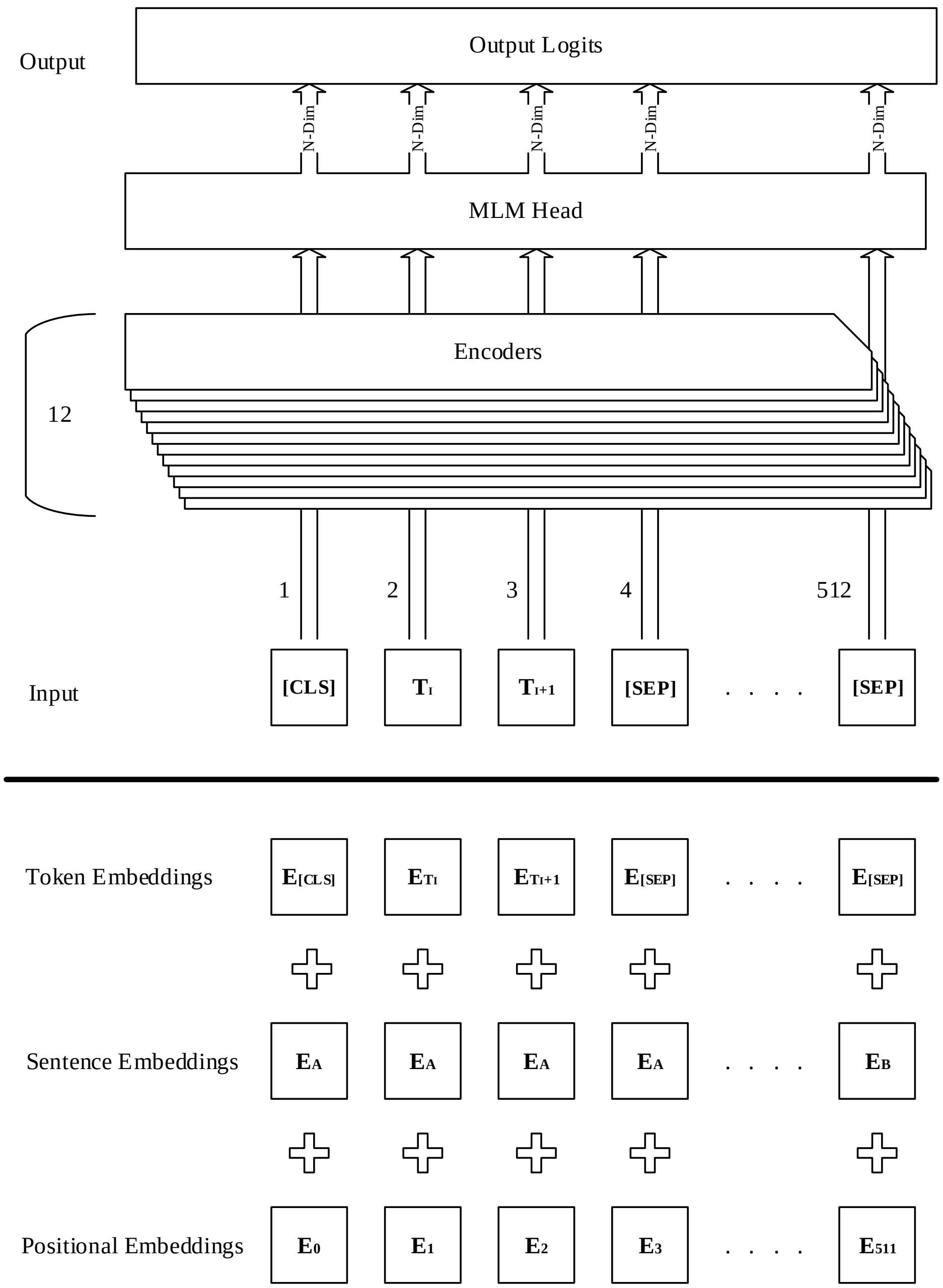

Figure 2. Architecture of PERCORE's Language Representation Model for Persian Language Spelling Correction Task.

### 3.4.5 Fine-tuning for Spelling Correction Task

We fine-tuned the language representation model on the Persian spelling correction task to achieve maximum performance for spelling correction. To fine-tune the model, we used TestSet1 from the Hamshahri corpus, which contains 103,840 sentences from the reserved articles. As input to the model, we used a single sentence that ended with a full stop. We only took one sentence at a time because our focus was on training the model for spelling correction. A closer examination of the test set indicated that several sentences were short, and masking a few tokens would eliminate a substantial amount of context. As a result, sentences with fewer than 20 words were excluded from the corpus. Finally, 91,420 sentences with a minimum length of 20 were chosen. However, because the input was a list of sentences that could not be directly fed into the model, we tokenized the text. The goal of the error correction task is to anticipate target or masked words by obtaining context from adjacent words. In other words, the model attempts to recreate the original sentence from the masked sentence received in the input at the output. Therefore, the target labels are the tokenizer's actual input_ids.

In the original $\mathbf{BERT_{BASE}}$ model, 15% of the input tokens were masked, with 80% replaced with [mask] tokens, 10% replaced with random tokens, and the remaining 10% left unchanged. However, in our fine-tuning task, we only replaced 15% of the input tokens with [mask], except for special ones; we did not use [mask] tokens to replace [SEP] and [CLS] tokens. We also avoided the random replacement of tokens in order to achieve better results. We employed a TensorFlow [78] background for training with Keras [79]. In addition, we used the Adam optimizer with a learning rate of $1\mathrm{E}-4$. The batch size was 32 and the model fully converged by the fourth epoch.

### 3.5 Persian Soundex

Soundex is a phonetic algorithm that indexes names based on their pronunciation in English [80]. It is widely recognized as one of the most popular phonetic algorithms. The primary objective of Soundex is to encode homophones with identical representations, allowing them to be matched despite minor variations in spelling.

In the case of the Persian language, there are 32 alphabets (in written form) that were divided into 14 categories based on their sounds. This classification is informed by extensive research in the field of Persian phonology [81, 82]. Our proposed code length for Persian is fixed at 4 characters. In our approach, Persian alphabet elements are grouped based on their pronunciation, with characters that have identical pronunciation being placed in the same classes, as shown in Table 2. When assigning codes, we ignored vowels. Some alphabets, such as "ا", "و" and "ی", can function as both a vowel and a consonant in Persian. When these letters appear at the beginning of a word, they act as consonants. As a result, syllabification is necessary to determine whether a letter is functioning as a vowel or a consonant.

Table 2. Soundex code for Persian Language Alphabet

| Set Number | Soundex Code | Characters with identical pronunciation |
|---|---|---|
| 1 | 0 | س ش ث ص |
| 2 | 1 | ط ت د |
| 3 | 2 | ذ ظ ض ز ژ |
| 4 | 3 | ج چ |
| 5 | 4 | ه ح ھ |
| 6 | 5 | ق ک خ |
| 7 | 6 | ب پ |
| 8 | 7 | گ غ |
| 9 | 8 | ا آ ع |
| 10 | 9 | م ن |
| 11 | A | ف |
| 12 | B | ل |
| 13 | C | و |
| 14 | D | ر |

In the evaluation section, we will examine the impact of the Persian Soundex algorithm on spelling correction accuracy. By analyzing the results, we can determine how effective the algorithm is in improving the accuracy of spelling correction for the Persian language.

### 3.6 Error Detection Module

The error detection module employs two distinct methodologies, depending on the type of error being detected. Lexical lookup is used for non-word errors, while contextual analysis is used for real-word errors. The first step in error detection, regardless of the type of error, is boundary detection and token identification. When the model receives a sentence $S$ as input, it first marks the beginning and end of the sentence with Beginning of Sentence ($BoS$) and End of Sentence ($EoS$) markers, respectively, and estimates the number of tokens in the sentence:

$$< BoS > T_1\ T_2\ T_3 \dots T_n < EoS >$$

It is important to note that the number of tokens is equal to the maximum number of iterations that the model will attempt to detect an error in the sentence.

#### 3.6.1 Non-word Error Detection

Spell checkers primarily utilize the lexical lookup approach to identify spelling errors. This method involves real-time comparison of each word in the input sentence against a reference dictionary, which is typically constructed using a hash table. Starting with the $BoS$ marker, every token in the sentence is scrutinized to determine its correctness based on word order. This procedure continues until the $EoS$ marker is encountered. However, if a word is detected as misspelled, the error detection cycle terminates and the error correction phase is initiated. The following is an example of non-word error detection:

مصرف مواد غذایی در جهان با زوند [روند]
سریع‌تری نسبت به تولید آن جریان دارد.

Food consumption in the world is growing faster than its production.

In the provided example, the intended ("روند" / rævænd / 'trend') was erroneously typed as 'زوند'. This error is a result of a substitution operation and is just one unit of distance away from the correct word. The model was able to efficiently detect this error.

#### 3.6.2 Real-word error detection

In this work, contextual analysis is employed for the identification of real-word errors. Traditional statistical models utilized n-gram language models to examine the frequency of a word's occurrence and assess the word's context by considering the frequency of the word appearing with "***n***" preceding terms. More modern approaches employ neural embeddings to evaluate the semantic fit of words within a given sentence. In our proposed method, we utilize the mask feature and leverage contextual scores derived from fine-tuned bidirectional language model to detect and correct word errors. The real-word error detection is explained as follows:

1- First, starting with the $BoS$ marker, the model attempts to encode each word as a masked word, beginning with the first word.
2- Second, a list of probable candidates for the masked word is obtained from the language representation model's output.
3- Thirdly, based on the error generation scenario, replacement candidates are generated within edit-distances of 1 and 2 from the masked word.
4- List of candidates, along with the original token, is checked against the language model's output for the masked token.
5- If any candidate has a higher probability value than the masked word, we consider the original word to be an error, and thus, the process ends. However, if no error is found, the model shifts one unit to the right, and the same steps are repeated for all words within the sentence until the $EoS$ marker is reached.

As mentioned earlier, based on our error generation strategy, a sentence may contain only one error. Therefore, as soon as an error is detected, the correction process commences immediately; afterward, the model proceeds to the next sentence. Pseudocode2 shows the algorithm for real-word error detection.

Pseudocode2: Real-word Error Detection Algorithm

| | |
|---|---|
| **Input:** | Persian text corpus as $P$ |
| **Output:** | Marked real-word errors |
| **Start:** | |
| **1** | Recognize all sentences $S_0, S_2, \ldots S_N$ boundaries |
| **2** | For each $S_i \in S_N$ do:<br>For all the words $W_0, W_1 \ldots W_N \in S_i$ do:<br>Mask the current word as $W_i$;<br>Get contextually similar tokens $T_0, T_1 \ldots, T_n$ for $W_i$ from the language model output $LMO$;<br>Generate replacement candidates $C_0, C_1 \ldots C_n$ within edit-distances of 1 and 2 from $W_i$; |
| | For each $C_i \in C_0, C_1 \ldots C_n$ do:<br><br>For each $T_J \in LMO$ do:<br><br>if $C_i == T_J$:<br>store $C_i$ AND context-score($C_i$) in rep_list;<br><br>For each $C_i \in$ rep_list do:<br>if context-score($C_i$) > context-score($W_i$):<br>mark $W_i$ as an error;<br>break; |

Here’s an example of successful real-world error detection:

[masked token]

توجه داشته باشید که روش های ارتباط **زمان** [زنان] با مردان متفاوت است.

Be aware that women's communication methods are different from men's

In the example provided, the term ( “زمان” /zæmɑn/ ‘time’ ) is identified as a real-word error. The user’s intended word was (“زنان” /zænɑn/ ‘women’ ). The model first encodes the masked token and inputs it into language representation model, which then produce a list of contextually appropriate tokens.

Subsequently, a list of candidate replacements is generated using the Damerau-Levenshtein distance measure. In this case, the edit-distance is 1. The model then compares the context similarity score of each replacement candidate with the output list from the language model.

Table 3 displays the context similarity scores of the top five replacement candidates from the output of the language representation model.

Table 3. Contextual Scores of the Top Five Replacement Candidates

| # | Replacement candidate | Contextual Score |
|---|---|---|
| 1 | زنان | 0.530 |
| 2 | زمان | 0.090 |
| 3 | زبان | 0.190 |
| 4 | رمان | 0.001 |
| 5 | زیان | 0.003 |

### 3.7 Error Correction Module

The error correction process is initiated when a misspelling is identified in the input. In this stage, a ranking algorithm is designed that mainly relies on the contextual scores from fine-tuned language representation model and phonetic similarity algorithm.

#### 3.7.1 Non-word Error Correction Process

In the process of non-word error correction, following steps are taken:

1- The model first utilizes the Damerau-Levenshtein edit distance measure to generate a set of replacement candidates within 1 or 2 edits.
2- The misspelled word is then encoded as a "mask" and input into the fine-tuned model.
3- The model retrieves all probable words from the output and matches them against the candidate list.
4- Next, the model retains a certain number of candidates with the highest contextual scores. Based on our observations, the optimal number is 10.
5- The method then compares the Soundex similarity between the erroneous word and remaining replacement candidates. If the error and candidate share the same code, that candidate is deemed the most suitable word. However, if two or more probable candidates carry the same Soundex code as the erroneous word, then the candidate with the highest contextual score is selected as the replacement for the error.

#### 3.7.2 Real-word Error Correction Process

In the case of real-word error correction, this process follows the real-word error detection process:

1- The contextual scores of probable candidates are retrieved from fine-tuned model.
2- The model stores a number of desired candidates with the highest contextual score. Based on our observations, the optimal number is 10.
3- The method compares the Soundex similarity between the word error and replacement candidates. If the error and the candidate share the same code, then that candidate is the most suitable word.
4- However, if two or more probable candidates carry the same Soundex code as the word error, then the candidate with the highest contextual score is chosen as the replacement for the error.

## 4. Evaluation and Results

In this section, we first examine the effect of fine-tuning various parameters on the performance of our proposed model. We then evaluate and compare the performance of our method against various baseline models in the spelling correction task. This will provide insight into the effectiveness and accuracy of our approach in detecting and correcting spelling errors.

### 4.1 Dataset

Our evaluation datasets comprise 94,379 reserved articles from the Hamshahri corpus. We collected articles from the twelve most frequently referred categories, including international, religious, economic, political, social, sports, literary, scientific, general, incidents, legal, and national security, into four unique datasets. Table 4 shows the specifics of each dataset.

Table 4. Dataset details

| Dataset name | TestSet1 | TestSet2 | TestSet3 | TestSet4 |
|---|---|---|---|---|
| number of articles | 15,712 | 6,204 | 3,455 | 69,008 |
| number of sentences | 103,840 | 86,035 | 57,638 | 123,512 |
| number of tokens | 3,496,720 | 3,191,898 | 1,700,321 | 6,546,136 |
| number of distinct tokens | 147,851 | 155,964 | 160,912 | 183,473 |

Our evaluation datasets are comprised of four different test sets from the Hamshahri corpus. TestSet1 contains 103,840 sentences from eight different genres: social, economic, law and national security, international, religious, sports, science, and politics. TestSet2 covers six various news categories and includes 155,964 distinct tokens. TestSet3 mainly includes 3,455 articles from five different genres. Finally, TestSet4 is comprised of 6,546,136 distinct tokens that cover eleven different genres.

### 4.2 Evaluation metrics

The primary evaluation measures for assessing the performance of models on non-word and real-word error detection and correction tasks are precision (P), recall (R), and F-measure (F1-Score). Precision (P) measures the accuracy of a model, while recall measures its exhaustiveness or sensitivity. The F1-Score, which is the weighted harmonic mean of both metrics, can be calculated by combining them. In F1, both precision and recall are given equal weight. Equation 5 describes the F1-Score evaluation measure.

$$F1 - Score = 2 * \frac{P * R}{P + R} \quad (5)$$

### 4.3 Experiments and Fine-tuning System Parameters

The experiments are divided into two main groups. The first set of experiments uses TestSet1, while the final set uses TestSet2, TestSet3, and TestSet4. The goal of the first experiment is to fine-tune the language representation model for spelling correction tasks, determine the impact of various error densities, and examine the effect of different edit-distance function values on the performance of word-error detection. The second series of experiments, on the other hand, investigates the effectiveness of the proposed model for detecting and correcting various types of errors and provides a meaningful comparison with other baselines

#### 4.3.1 Fine-tuning Error Generation Algorithm

We investigate the impact of various error densities and edit-distance function values on the error-detection task. To do this, we use the TestSet1 corpus to examine how these parameters affect the context-sensitive error detection performance of the model. We first inject context errors into the sample text and then assess context-sensitive error detection using the erroneous text. Similar strategies have been employed in previous studies [1, 15, 16, 34, 63]. To build the erroneous corpora, we randomly selected 10,000 sentences from the TestSet1 corpus and populated them with context-sensitive errors using Pseudocode1.

Authors in [8] reported that 80% of word errors are within distance 1, and 20% are within distance 2. We used the same values in the error-generation algorithm ($D1$=0.8). In addition, based on observations from [34, 63], average values of 0.5 to 0.85 for $\beta_1$and 0.16 to 0.20 for $\beta_2$ are suitable. However, since we are evaluating the real-word error detection performance, we assign values of 0.0 and 1.0 to the aforementioned variables (($\beta_1$=0.0 $\beta_2$=1.0), which means all the generated errors are real-word errors. For the edit distance, we used Damerau-Levenshtein since it treats the swapping of two adjacent letters as a single operation, whereas Levenshtein requires two operations. We generate context errors that are within 1 and 2 edit-distance of the original word using the Damerau-Levenshtein measure $dDL$ = (1,2). Another factor we must consider is the number of errors. We used various $E$ values between 10% to 50% to check the performance of the model. In other words, if the "one-error-per-sentence" rule is strictly adhered to, 10% to 50% of the sentences will contain an error. As a result, when $E$ equals 10%, $D1$ equals 0.0, $\beta_1$ equals 0.0, and $\beta_2$ equals 1, the total number of errors will be 1000. This amounts to zero non-word errors and 1000 real-word errors, where 800 real-word errors are within edit-distance 1 and 200 are within distance 2.

To test the resilience of our method and simulate various error sources, we employ error densities of 10%, 20%, 30%, 40%, and 50%, where 50% represents a highly erroneous source. We want to assess the accuracy of our model and see how different error densities and edit-distance function values affect the error detection success rate. Table 5 summarizes the results on the TestSet1 corpus in terms of $F1 - Score$. The results indicate that the highest $F1 - Score$ values are attained when the error density is set to a minimum value of 10%. In this scenario, the overall detection $F1 - Score$ is 0.883, and the difference between the detection $F1 - Scores$ for edit-distance 1 and edit-distance 2 is 3.8%. The reason for this difference is that as the number of replacement candidates in higher edit-distances increases, it becomes more likely for the model to mistakenly identify a correct word as an error.

As indicated in Table 5, there is a gradual decrease in the system's overall $F1 - Score$ as the value of $E$ increases. However, for higher error density values such as 40% and 50%, the numbers are more stable. When the error density is set to 40%, the $F1 - Score$ begins to converge. In this situation, the detection F1-Scores for detecting errors at distances 1 and 2 are 0.836 and 0.803 respectively, while the overall $F1 - Score$ is 0.829. These results validate that our proposed approach is highly accurate and can effectively identify real-word errors across various distances, even when the frequency of errors in a given corpus is extremely high.

We also carried out an in-depth examination of the mistakes made by our model. We observed that the model tended to overlook errors when the artificially introduced error had a semantic connection to the context words. For example, in the original word sequence " جهت اصلاح مسیل" (in order to repair a stream), the error generation algorithm substituted the word ("مسیل" /mæsil/ 'stream' ) with the artificially introduced error ("مسیر" /mæsir/ 'path' ) , which is within edit distance 1. This led to the word sequence "جهت اصلاح مسیر" (in order to correct a path), which had a higher context similarity score than the original word sequence. As a result, this word sequence was overlooked by the model.

While this issue has not been mentioned in prior research on Persian spelling correction, we consider it to be a major challenge when introducing artificially generated errors into Persian corpora. This is probably due to the absence of comprehensive and authentic error corpora for the Persian language. Nonetheless, this problem can be addressed by verifying the generated errors against a list of N-grams and contextual similarity scores within the error generation algorithm.

Table 5. The Proposed approach's performance for various values of $E$ and edit-distance function on real-word error detection task

| $E$ | Model | edit-distance1 | edit-distance2 | overall result |
|---|---|---|---|---|
| 10% | PERCORE | 0.891 | 0.853 | 0.883 |
| 20% | PERCORE | 0.866 | 0.829 | 0.859 |
| 30% | PERCORE | 0.848 | 0.813 | 0.841 |
| 40% | PERCORE | 0.836 | 0.803 | 0.829 |
| 50% | PERCORE | 0.832 | 0.800 | 0.826 |

### 4.3.3 Preparing Datasets for Full Evaluation

At this stage, we are preparing erroneous text to evaluate the performance of our proposed model and other baseline methods in both real-word and non-word error detection and correction tasks. Initially, we randomly selected 10,000 sentences from each of the TestSet2, TestSet3, and TestSet4 corpora, resulting in a total of 30,000 sentences. We then used Pseudocode1 to generate the erroneous corpora. The default configuration for error generation includes the following parameter values:

1) $N$ = 10,000; 2) $E$ =10%,50%; 3) $D1$=0.8; 4) $\beta_1$=0.8; 5) $\beta_2$=0.2;

Based on the error generation settings, 30,000 sentences were randomly selected from all three datasets to create the real-word and non-word error test set. The Damerau-Levenshtein edit-distance of 1 or 2 was applied to the target sets to generate artificial errors resulting from insertion, deletion, transposition, or substitution operations. Of the 30,000 sentences, 24,000 included an error within edit-distance 1, where 19,200 were non-word errors and 4,800 were real-word errors. Additionally, 6,000 sentences contained errors within edit-distance 2; of these, 4,800 were spelling errors and the remainder were real-word errors.

## 4.4 Baseline Models

In our research, we implemented several baseline models for non-word and real-word error correction tasks to ensure a fair comparison. All models, including Perspell by Dastgheib et al. [34], the four-gram model [41], and a Persian Continuous Bag-of-Words (CBOW) model [83], were developed using Python and trained on the same dataset as PERCORE.

For real-word error correction, we replicated the methodology of two distinguished models. The first is based on confusion sets, as proposed by Dastgheib et al. [34]. The second is the real-word error correction module from the Vafa spell-checker [63], a tool widely used for identifying and correcting real-word errors in Persian texts. By comparing these models, we aim to understand their strengths and weaknesses, and leverage this understanding to enhance error correction in Persian language processing.

### 4.4.1 Perspell

Perspell, a statistical spelling correction framework tailored for Persian, exploits a dictionary lookup strategy to pinpoint non-word errors. [34]. Leveraging a bigram language model, Perspell discerns the most contextually fitting candidate for correction. It differentiates itself by employing predefined confusion sets and harnessing the rich lexicon of Persian WordNet to extract synonyms, thereby enhancing real-word error detection.

Adaptations for auto-correction have seen the introduction of an advanced bigram language model that meticulously combs through related bigrams for each candidate. The enhancement involves sorting candidates based on probability values, ensuring the selection of the most probable candidate for correction. This strategic approach empowers Perspell to offer corrections, significantly reducing the manual review workload.

### 4.4.2 Yazdani, et al. Model

The Yazdani et al. model emerges as a benchmark in Persian non-word error correction through its utilization of a weighted bi-directional fourgram language model. [41]. This model's core lies in its application of a nuanced quadripartite equation, designed to accord precedence to n-grams based on their sequential order.

This methodical prioritization facilitates a nuanced and highly accurate error correction process, showcasing the model's unparalleled effectiveness in refining Persian texts. The innovative use of a bi-directional approach allows for a more holistic analysis of text, considering both preceding and succeeding context to identify the optimal replacement, thereby setting a new standard in language processing precision.

### 4.4.3 CBOW Model

The Continuous Bag of Words (CBOW) model represents a leap forward in understanding word meanings through contextual analysis [83]. Focused on predicting suitable words within specific contexts, the CBOW architecture is instrumental in identifying target words amidst source context words. Its training on a corpus of 1.4 million documents from PERCORE signifies a substantial effort to refine its prediction capabilities. With technical parameters such as a context window size of 10 and a dimension size of 300, the model utilizes input and output matrices to calculate the hidden layer effectively. This extensive training empowers the CBOW model to excel in non-word error correction tasks, showcasing its robustness in language modeling and its potential for broader linguistic applications.

### 4.4.4 Vafa Spell-checker

The Vafa spell-checker, specifically engineered for Persian text, adopts a comprehensive three-step methodology to tackle real-word errors [63]. It initiates the process with a detailed contextual analysis, considering adjacent words to gauge the context accurately. The model then generates a list of potential replacements, contemplating every conceivable single-letter modification and semantically akin words. This process is further refined using predefined confusion sets, enhancing the model's capability to detect real-word errors with high precision. Employing a trigram language model, the Vafa spell-checker not only identifies but also corrects errors, leveraging its sophisticated algorithm to ensure textual integrity. This model's implementation highlights its effectiveness in enhancing the quality of Persian texts, providing a valuable tool for language practitioners.

### 4.4.5 GPT-2.0 for Persian Spelling Correction

GPT-2.0 marks a significant milestone in NLP, acclaimed for its generative prowess and profound understanding of context [84]. Its pre-training, conducted on a varied array of datasets, lays a solid foundation for fine-tuning endeavors, including the delicate task of spelling correction in Persian. Importantly, the initial pre-training of GPT-2.0 leveraged an extensive corpus of 1.4 million documents to ensure comprehensive coverage of Persian's linguistic diversity, providing a robust starting point for fine-tuning related to spelling correction.

- **GPT-2.0 Implementation and Fine-tuning**

*Approach:*

- **Model Preparation:** We initiate the process with a GPT-2.0 model pre-trained on a diverse dataset, including a significant corpus of 1.4 million documents to immerse the model in Persian language intricacies. This pre-training phase is crucial for acquainting the model with the nuances of Persian, setting the stage for its subsequent fine-tuning on the TestSet1 corpus.
- **Fine-tuning Strategy:** Utilizing a corpus of 91,420 sentences from TestSet1 of the Hamshahri corpus, each with a minimum of 20 words and infused with artificially generated spelling errors, effectively simulates real-world spelling correction challenges. This targeted exposure is crucial for the model to acquire and refine strategies for accurately identifying and correcting errors in Persian texts.

*Hyperparameters*:

- **Learning Rate:** A carefully chosen learning rate of 5e-5 supports gradual and precise model adjustments, enhancing its performance in spelling correction tasks.
- **Batch Size:** Opting for a batch size of 16 strikes a balance between computational efficiency and the richness of linguistic input.
- **Epochs:** Limiting the training to 4 epochs helps avoid overfitting, fostering a model that remains versatile across different contexts.

### 4.4.6 GPT-3.0 for Persian Spelling Correction

GPT-3.0, as the successor to GPT-2.0, elevates the capabilities of NLP models through its vast scale and intricate architecture [21]. By undergoing pre-training on an even broader spectrum of data, including the aforementioned 1.4 million documents to enrich its understanding of Persian, GPT-3.0 showcases unparalleled adaptability to a wide range of NLP tasks. Its precision in generating text that is both contextually relevant and grammatically coherent positions it as an indispensable asset for refining spelling correction techniques, particularly within the complex linguistic framework of Persian.

- **GPT-3.0 Implementation and Fine-tuning**

*Approach:*

- **API Utilization:** We harness GPT-3.0's capabilities through OpenAI's API, creating a dynamic environment where sentences with intentional spelling errors are processed, mirroring the complexities of Persian spelling correction.
- **Prompt Engineering:** A critical component of methodology is the crafting of prompts that mirror the linguistic diversity and commonality of spelling mistakes within Persian. By leveraging a comprehensive corpus of 1.4 million documents, we generate prompts that blend correctly spelled words with deliberately introduced spelling errors. This rich dataset serves to acquaint GPT-3.0 with the wide array of spelling inaccuracies characteristic of Persian, guiding the model towards making accurate corrections based on contextual clues. For fine-tuning, we specifically utilize 91,420 sentences from TestSet1 of the Hamshahri corpus, each with a minimum of 20 words, to ensure the model is finely attuned to the task of correcting spelling errors in Persian texts.

*Hyperparameters (via API settings):*

- **Temperature:** The temperature is carefully adjusted to 0.7, optimizing the model's output for a judicious mix of creativity and precision. This setting is pivotal in ensuring that the corrections made by GPT-3.0 are not only imaginative but also adhere strictly to Persian orthographic standards.
- **Max Tokens:** With the average length of sentences in our training corpus at 23 words and the maximum length at 90 words, we adjust the Max Tokens setting accordingly. This calibration ensures that GPT-3.0 can fully process and correct errors within sentences of these lengths, providing contextually appropriate corrections that accommodate the variability and complexity characteristic of the Persian language.
- **Max Tokens and Top P:** Set at 0.95, enabling the model to consider a broad range of correction possibilities while prioritizing those most likely to be accurate within the specific context of each sentence.

## 5. Results and Analysis

In this section, we present the evaluations and results of both the proposed model and the baselines. To assess the effectiveness of our Persian Soundex algorithm, we employed two distinct correction strategies. The first strategy involves ranking replacements and correcting errors based solely on contextual scores. The second strategy involves applying a comprehensive correction method. These strategies provide a thorough evaluation of our model's performance in comparison to the baselines.

### 5.1 Non-word Error Correction Evaluation

In the first stage of evaluation, we compare the performance of our proposed approach to that of the aforementioned baseline models in terms of non-word error correction. It is important to note that since all the models utilize a dictionary look-up approach for detecting misspellings, the F1-score for misspelling detection is 100%. Table 6 presents the results of the non-word error correction task, providing a detailed comparison of the effectiveness of our approach and the baseline models. We evaluate all the models in two different scenarios: first, with an error density of 10% that simulates a normal source of errors; and second, with an error density set to 50%, which represents a very noisy source.

Table 6. Comparison of Various models' Performance on Non-word Error Correction Task

| **Model** | $E$ | edit-distance1 | edit-distance2 | overall result |
|---|---|---|---|---|
| PERCORE | 10% | 0.886 | 0.849 | 0.879 |
| PERCORE + Soundex | 10% | 0.898 | 0.863 | **0.891** |
| Perspell | 10% | 0.614 | 0.568 | 0.605 |
| Yazdani, et al. | 10% | 0.716 | 0.674 | 0.708 |
| Continuous Bag-of-Words (CBOW) | 10% | 0.752 | 0.701 | 0.742 |
| GPT-2.0 | 10% | 0.818 | 0.776 | 0.810 |
| GPT-3.0 | 10% | 0.872 | 0.834 | 0.864 |
| PERCORE | 50% | 0.838 | 0.813 | 0.833 |
| PERCORE + Soundex | 50% | 0.850 | 0.824 | **0.845** |
| Perspell | 50% | 0.559 | 0.515 | 0.550 |
| Yazdani, et al. | 50% | 0.660 | 0.619 | 0.652 |
| Continuous Bag-of-Words (CBOW) | 50% | 0.703 | 0.653 | 0.693 |
| GPT-2.0 | 50% | 0.798 | 0.756 | 0.788 |
| GPT-3.0 | 50% | 0.850 | 0.812 | 0.842 |

Table 6 provides a comprehensive comparison of the performance of various models on the non-word error correction task. Two configurations of PERCORE are compared with statistical baselines and the CBOW model. The results clearly show that both configurations of PERCORE outperform the other models, demonstrating superior performance and stability across different levels of error density. When the error density ($E$) is set to 10% and the Soundex algorithm is employed, PERCORE achieves its best performance with an $F1-Score$ of 0.891. This model effectively corrects misspellings within both edit-distance 1 and 2. The combination of contextual similarity with the Soundex algorithm proves to be the most robust scheme, offering a 1.2% increase in correcting non-word errors compared to using only contextual scores.

In contrast, Perspell shows the lowest performance, with $F1-Score$ of 0.605 and 0.550 for different error density values. The Contextual Scores + Soundex scheme outperforms Yazdani et al.'s approach by 18.3% when ($E$) equals 10%, and by 19.3% when E is set to 50%. These results demonstrate the robustness of the proposed approach even when the frequency of errors in the given corpora is high. The CBOW baseline was also significantly outperformed. According to Table 6, Perspell's Bigram method yielded the lowest $F1-Score$. However, using n-grams of higher order in Yazdani et al.'s approach appears to be effective. The CBOW model offers significantly better performance than the statistical models.

GPT-2.0 and GPT-3.0 show strong performance in non-word error correction tasks, with GPT-3.0 outperforming GPT-2.0 at both 10% and 50% error densities. This indicates the advancements in model capabilities and training methodologies from GPT-2.0 to GPT-3.0, showcasing improved context understanding and error correction effectiveness. However, the proposed PERCORE + Soundex approach outperforms both GPT models, particularly at a 10% error density, highlighting the tailored effectiveness of PERCORE when combined with phonetic analysis through Soundex for the Persian language.

In terms of PERCORE, the results of the Contextual Scores + Soundex scheme are notably better than those achieved using only Contextual scores. The best results are achieved when the pretrained model is used in conjunction with the Soundex phonetic similarity algorithm. Observations confirm that Soundex significantly improves results as substitution errors account for 41.7% of all errors in the test corpus when compared to insertion, deletion, and transposition errors, and most substitution errors are either phonetically or visually similar.

### 5.2 Real-word Error Detection and Correction Evaluation

We conducted an exhaustive comparison between our proposed model and the established baselines, focusing on the detection and correction of real-word errors in the Persian language. The results of our evaluations are presented in Table 7.

In the task of real-word error detection, our proposed approach outperformed both baselines. Our model, PERCORE, achieved its best performance when the error density ($E$) was set to 10%. In this scenario, the test corpora contained 600 test instances at distances 1 and 2, yielding an overall $F1 - Score$ of 0.890. The method proved to be robust, with a mere 4.7% difference in the overall $F1 - Score$ between the largest ($E$ = 50%) and smallest ($E$ = 10%) error density values. Furthermore, it demonstrated promising performance in detecting errors at both distances 1 and 2.

The authors of [34] claimed to have achieved an $F1 - Score$ of 0.726 for real-word error correction; however, we were unable to replicate this result in our evaluations. We subjected all models to a large number of test instances to evaluate their practical performance. In the case of Perspell, it achieved a maximum $F1 - Score$ of 0.653 for detecting real-word errors with an error density of 10% ($E$ = 10%). However, Perspell encountered difficulties in detecting real-word errors within distance 2 due to its exclusive reliance on confusion sets and the presence of a large number of potential candidates. This suggests that Perspell's effectiveness in detecting errors at distance 2 is somewhat limited. In our tests, Vafa Spell-checker delivered the least impressive results, with a maximum $F1 - Score$ of just 0.182 when the error density was set to 10%. Our findings suggest that it struggled to detect real-word errors at both distances 1 and 2. Performance of GPT models is competitive but yet behind PERCORE, with GPT-3.0 achieving an F1-Score of 0.871 and GPT-2.0 at 0.805 for the same error density. This highlights the strength of PERCORE in leveraging context and phonetic similarity for error detection, matching closely with GPT-3.0's advanced contextual understanding capabilities.

Table 7. Performance Evaluation on Real-word Error Correction Task

| Task | Model | $E$ | edit-distance1 | edit-distance2 | overall result |
|---|---|---|---|---|---|
| ***Real-word Error Detection:*** | | | | | |
| | PERCORE | 10% | 0.897 | 0.860 | **0.890** |
| | Perspell | 10% | 0.668 | 0.593 | 0.653 |
| | Vafa Spell-checker | 10% | 0.188 | 0.159 | 0.182 |
| | GPT-2.0 | 10% | 0.813 | 0.765 | 0.805 |
| | GPT-3.0 | 10% | 0.879 | 0.831 | 0.871 |
| | PERCORE | 50% | 0.850 | 0.814 | **0.843** |
| | Perspell | 50% | 0.618 | 0.558 | 0.606 |
| | Vafa Spell-checker | 50% | 0.164 | 0.133 | 0.158 |
| | GPT-2.0 | 50% | 0.795 | 0.747 | 0.787 |
| | GPT-3.0 | 50% | 0.861 | 0.813 | 0.853 |
| ***Real-word Error Correction:*** | | | | | |
| | PERCORE | 10% | 0.900 | 0.861 | 0.892 |
| | PERCORE + Soundex | 10% | 0.913 | 0.873 | **0.905** |
| | Perspell | 10% | 0.349 | 0.322 | 0.344 |
| | Vafa Spell-checker | 10% | 0.315 | 0.290 | 0.310 |
| | GPT-2.0 | 10% | 0.818 | 0.774 | 0.810 |
| | GPT-3.0 | 10% | 0.901 | 0.859 | 0.886 |
| | PERCORE | 50% | 0.856 | 0.833 | 0.851 |
| | PERCORE + Soundex | 50% | 0.868 | 0.846 | **0.864** |
| | Perspell | 50% | 0.324 | 0.293 | 0.318 |
| | Vafa Spell-checker | 50% | 0.297 | 0.268 | 0.291 |
| | GPT-2.0 | 50% | 0.798 | 0.756 | 0.783 |
| | GPT-3.0 | 50% | 0.867 | 0.823 | 0.852 |

We evaluated the proficiency of all models in rectifying real-word errors. As shown in Table 7, our suggested method excels in correcting real-word errors at various distances and densities, outperforming the other models. The peak $F1 - Score$ of 0.905 is attained when the error density is set to 10% and the Persian Soundex algorithm is employed. Conversely, the lowest $F1 - Score$ of 0.851 is achieved with an error density of 50%, where the model relies solely on the contextual score from the pretrained model. The 5.4% difference is justifiable given the substantial number of corrections made. The results for correcting context-errors at distance 2 are also commendable, indicating that our suggested method is both robust and accurate in rectifying detected context-errors. GPT-3.0 recorded an F1-Score of 0.886 in real-word error correction, while GPT-2.0 scored 0.810. Although GPT-3.0 presents a formidable capability in generating contextually appropriate corrections, PERCORE + Soundex surpasses it, suggesting that for the specific challenges of Persian spelling correction, the incorporation of phonetic similarity provides an edge. The Perspell correction software attains its optimal $F1 - Score$ of 0.344 for correcting real-word errors when the error density is set to 10%. However, its performance significantly deteriorates when the error density escalates to 50%, leading to a 2.6% reduction in $F1 - Score$. This implies that Perspell often fails to substitute detected word errors with the correct candidate, particularly when correcting real-word errors at distance 2.

Among the six models, Vafa Spell-checker exhibits the least impressive performance, with a peak $F1 - Score$ of 0.310 achieved at an error density of 10%. Its performance further declines when correcting context-errors at distance 2.

In summary, the results of our study affirm the effectiveness of the proposed approach in both correcting non-word errors and accurately detecting and correcting real-word errors within Persian texts. Our approach demonstrates a notable improvement over traditional baseline models, offering a more nuanced understanding and handling of the Persian language's unique characteristics. By integrating contextual analysis with phonetic considerations, our system achieves a balanced performance that addresses common challenges in spelling correction tasks.

## 6. Discussion

This research has made considerable advancements in the field of spelling correction for the Persian language, specifically addressing the automatic detection and correction of two major spelling errors: non-word and real-word errors. A four-stage architecture was developed, which not only preprocesses and normalizes free-running Persian text but also generates error corpora with artificially induced errors of different types at the desired density level. This architecture is capable of auto-detecting and correcting real-word and non-word errors with high precision, marking a significant improvement over earlier Persian spelling correction models.

The proposed approach leverages a state-of-the-art language representation model fine-tuned for the Persian language spelling correction task. The model uses contextual scores for both error detection and correction. The integration of the developed Persian Soundex phonetic matching algorithm with the contextual score from the language model significantly increases the correction success rate. The robustness of our model was tested under various error densities, demonstrating its resilience even under high error density values. In order to make a fair comparison, three baselines were imitated and developed, all trained with the same high-quality, manageable samples as the language representation model. According to the evaluations, our method outperformed all the baselines with outstanding results, achieving F1-Scores of 0.890 and 0.905 in the detection and correction of real-word errors respectively. For non-word error correction, our model also exhibits promise with an F1-Score of 0.891.

While GPT models exhibit commendable performance, their proficiency in addressing the nuanced complexities of Persian spelling errors reveals inherent limitations. This observation highlights the superiority of incorporating specialized phonetic algorithms alongside contextual analysis within our proposed model. Specifically, GPT-2.0, despite its efficacy, demonstrates a noticeable shortfall in precisely detecting real-word errors compared to GPT-3.0. This discrepancy can be attributed to GPT-2.0's relatively smaller model size and its less sophisticated grasp of the intricate linguistic patterns that characterize Persian text.

The application of GPT-3.0, especially via its API service, necessitates a careful evaluation of cost and real-time usability concerns. The model's per-request pricing structure poses a potential financial burden on applications necessitating high-frequency interactions, thereby impacting developers with limited budgets and smaller entities. Moreover, despite efforts to minimize latency, the inherent delay in processing API requests might compromise user experience in scenarios where immediate feedback is essential.

To navigate these challenges, a hybrid methodology, leveraging lightweight, local models for initial processing supplemented by selective use of the GPT-3.0 API for more complex inquiries, emerges as a pragmatic solution. This approach not only seeks to reduce operational expenses but also to improve system responsiveness, facilitating a smoother interaction for users.

One clear limitation of our evaluations was that it is dedicated to detection and correction of non-word and real-word errors and may fail in handling grammatical errors. This is an area that could be explored in future research.

Another limitation encountered during this study was dealing with cases where our system was unable to identify the correct replacement word. After ranking the candidates based on contextual scores and Soundex code comparison, the model selects the top word in the ranked correction candidates list. However, this may not always align with the user's intended word.

For instance, consider the sentence “در راستای تحدید قرارداد” (in order to limit the contract). The system accurately identified the context error word, (“تحدید” /tæhdid/ ‘limit’). The correction is (“تمدید” /tæmdid/ ‘extend’), but among the most probable candidates are: (“تمدید” /tæmdid/ ‘extend’), (“تجدید” /tædʒdid/ ‘extend’) and (“تشدید” /tæʃdid/ ‘intensify’). The intended correction is second on the list, but due to close contextual scores of the first and second entries, either could be chosen as correct. The model also compares Soundex similarity of candidates to the error. Persian Soundex code for error is “1411”, while for candidate words(“تمدید” /tæmdid/ ‘extend’), (“تجدید” /tædʒdid/ ‘extend’) and (“تشدید” /tæʃdid/ ‘intensify’) it's “1911”, “1311” and “1011” respectively. Given that the difference between Soundex code for error and candidates is limited to one unit, the model leverages contextual score to select best correction candidate. As first candidate has higher contextual score, it replaces context error.

To mitigate this issue, we propose incorporating orthographic features of Persian alphabet into our model. This could be achieved by developing an exclusive Persian Shapex algorithm that considers orthography of letters. For instance, characters such as “ح“ ,”ج“ ,”چ”, and “خ” are very similar in terms of orthography, so we could group them together and assign them with same code. This methodology could help address problems in identification of probable candidates.

Furthermore, we found that the model was more likely to misdetect errors when the artificially generated error was semantically related to the context words. This is likely due to the lack of genuine error corpora in Persian. However, this issue could be mitigated by checking generated errors against the list of N-grams in the error generation algorithm.

The practical implications of our research findings suggest that our proposed spelling correction model could be implemented in real-world applications to enhance text processing systems in Persian language. This opens up new avenues for improving text-based communication systems, educational platforms, and digital content creation tools that cater to Persian language users.

## 7. Conclusions

In this study, we introduced a novel approach for detecting and correcting both non-word and real-word errors in Persian text. Our method, which leverages a state-of-the-art language representation model fine-tuned for the Persian language spelling correction task, has demonstrated superior performance compared to previous models. It achieved F1-Scores of 0.890 and 0.905 in the detection and correction of real-word errors respectively, and an F1-Score of 0.891 for non-word error correction.

Moreover, our approach is robust to variations in error density and dataset size, effectively handling a wide range of real-world errors. The integration of the developed Persian Soundex phonetic matching algorithm with the contextual score from the pre-trained model significantly increases the correction success rate.

We believe that our method represents a significant advancement in the field of spelling error detection and correction for Persian text. By improving the quality of Persian text on the internet and other digital media, our approach has the potential to make a meaningful impact on the Persian-speaking community.

In our future work, we aim to leverage the capabilities of large language models to address the complex challenge of correcting multiple errors within a single sentence. By integrating the structured semantic framework of ontologies like FarsNet with the dynamic, contextual prowess of large language models, we plan to significantly enhance our model's precision and adaptability. This approach, complemented by the incorporation of orthographic similarity features through the Persian Shapex algorithm and new strategies for grammatical error correction, aims to advance the robustness and applicability of our method in Persian language processing.

### Data Availability

The dataset supporting this article are from previously reported studies and datasets (Hamshahri corpus), which have been cited. The data are available at:

https://dbrg.ut.ac.ir/hamshahri/

### Conflicts of Interests

The author(s) declare(s) that there is no conflict of interest regarding the publication of this paper

### Funding Statement

The research did not receive any specific funding

# Declarations

### Ethical Approval

Not Applicable

### Authors' contributions

All authors have made substantial contributions to the conception and design of the research, as well as the drafting and reviewing the manuscript.

### Acknowledgments

Not Applicable